\documentclass[a4paper]{article}

\usepackage{INTERSPEECH2022}
\usepackage{multirow}
\title{Reducing Multilingual Context Confusion for End-to-end Code-switching Automatic Speech Recognition}
\name{Shuai Zhang$^{1,2}$, Jiangyan Yi$^{2}$, Zhengkun Tian$^{1,2}$, Jianhua Tao$^{1,2,3}$, Yu Ting Yeung$^{4}$, Liqun Deng$^{4}$}
\address{
	$^1$School of Artificial Intelligence, University of Chinese Academy of Sciences, China \\
	$^2$NLPR, Institute of Automation, Chinese Academy of Sciences, China \\
	$^3$CAS Center for Excellence in Brain Science and Intelligence Technology, China\\
	$^4$Huawei Noah's Ark Lab, Shenzhen, China
}
\email{\{shuai.zhang, jiangyan.yi, zhengkun.tian, jhtao\}@nlpr.ia.ac.cn,\\
	\{yeung.yu.ting,  dengliqun.deng\}@huawei.com}

\begin{document}
	\maketitle
	\begin{abstract}
		Code-switching deals with alternative languages in communication process. Training end-to-end (E2E) automatic speech recognition (ASR) systems for code-switching is especially challenging as code-switching training data are always insufficient to combat
		the increased multilingual context confusion due to the presence of more than one language. We propose a language-related attention mechanism to reduce multilingual context confusion for the E2E code-switching ASR model based on the Equivalence Constraint (EC) Theory. The linguistic theory requires that any monolingual fragment that occurs in the code-switching sentence must occur in one of the monolingual sentences. The theory establishes a bridge between monolingual data and code-switching data. We leverage this linguistics theory to design the code-switching E2E ASR model. The proposed model efficiently transfers language knowledge from rich monolingual data to improve the performance of the code-switching ASR model. We evaluate our model on ASRU 2019 Mandarin-English code-switching challenge dataset. Compared to the baseline model, our proposed model achieves a 17.12\% relative error reduction. 
	\end{abstract}
	
	\section{Introduction}
	Code-switching refers to the use of multiple languages in communication, which is a common language phenomenon especially for bilingual speakers \cite{muysken2000bilingual}.  
	E2E ASR models achieve excellent performance on monolingual speech recognition with joint optimization of acoustic, pronunciation, and language models \cite{graves2006connectionist,graves2013speech,chan2016listen,8462506,rao2017exploring,gulati2020conformer}. 
	Training an E2E code-switching ASR system is still challenging as it heavily relies on audio-text data pairs \cite{li2019towards,8682674,8462201,zhang2021decouple}.  The presence of more than one language in an utterance increases the chance of multilingual context confusion. 
	
	To alleviate the problem of multilingual context confusion, a natural idea is to use a large amount of monolingual audio-text data to train the code-switching ASR model. Although large datasets significantly improve the performance of monolingual ASR tasks with E2E models \cite{graves2013speech,chan2016listen,8462506}, it is still insufficient to handle code-switching problem with conventional E2E models.
	A possible reason is that multilingual context information is not aligned from monolingual data of different languages during model training. Code-switching text data augmentation technology tries to alleviate the multilingual context confusion problem. The method often applies a translation system to artificially generate code-switching text from monolingual text according to various rules \cite{DBLP:conf/interspeech/YilmazHL18,DBLP:conf/interspeech/ChangCL19}. The generated text data increase the richness of multilingual context, and improve the system's ability to model knowledge of different languages. To generate more reasonable code-switching text, various linguistic theories are proposed to constrain the generation strategy \cite{li2012code,li2013improved,li2014language}. There are many attempts to explain the grammatical constraints on code-switching. The three most widely accepted theories including the Embedded Matrix (EM) \cite{joshi1982processing}, the Equivalence Constraint (EC) \cite{pfaff1979constraints} and the Functional Head Constraint (FHC) theories \cite{bhatt1995code}. Although applying linguistic theories to text data generation improves the performance of code-switching ASR, text data only assist E2E ASR models indirectly in the form of language models \cite{DBLP:conf/interspeech/SriramJSC18,DBLP:conf/interspeech/ZhaoSRRBLP19,DBLP:conf/interspeech/BaiYTTW19}. Moreover, text data generation requires extra computational cost and increases the complexity of ASR training pipeline. Therefore, there is great benefit if we can directly apply linguistic theories to an E2E code-switching ASR model. 
	
	\begin{figure*}[htb]
		
		\begin{minipage}[b]{1.0\linewidth}
			\centering
			\centerline{\includegraphics[width=16.0cm]{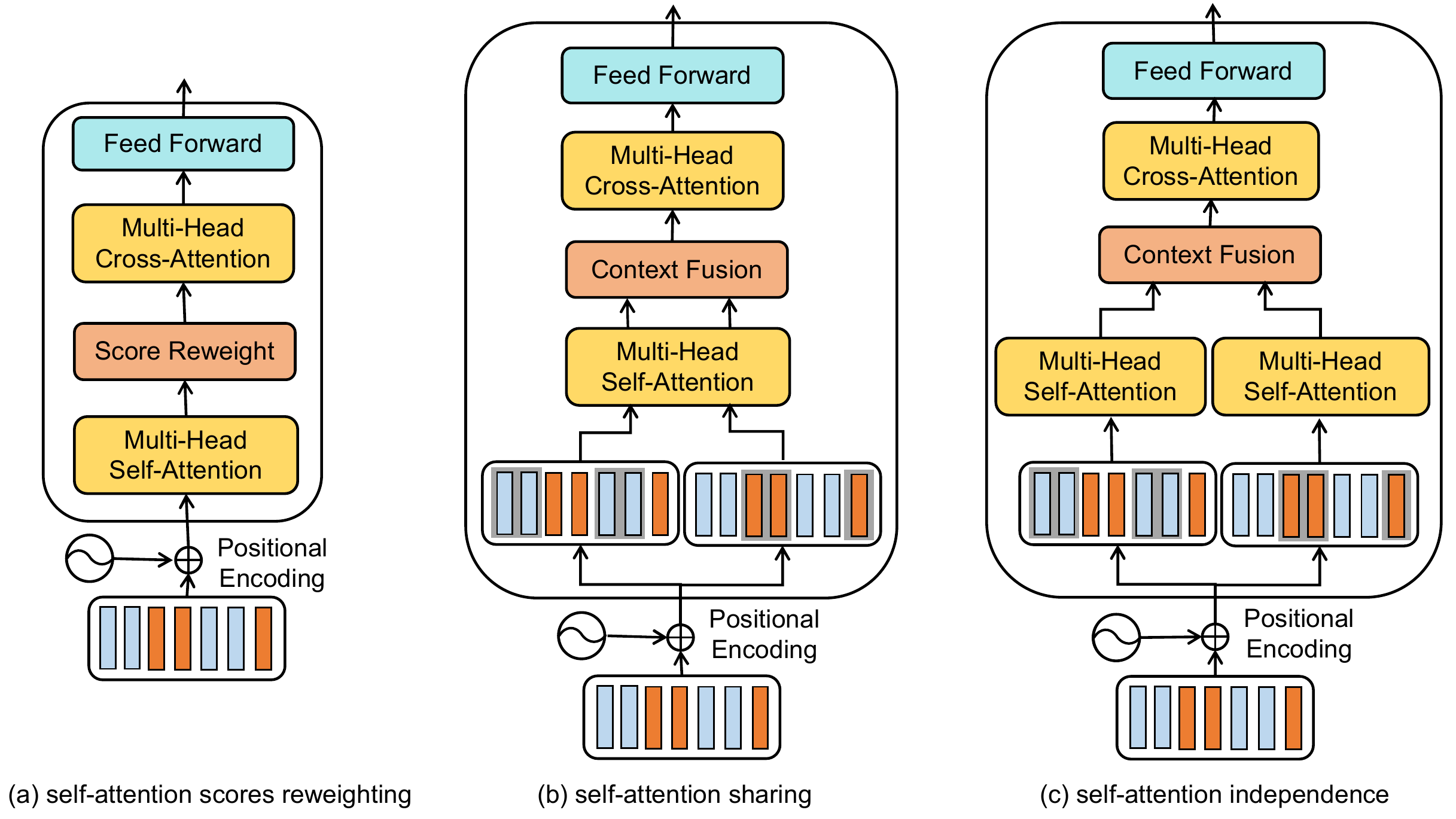}}
			\centerline{}
			\caption{Three specific language-related attention mechanisms of our proposed method. (a) Adjust the weights of the self-attention scores according to language. (b) The embeddings of the two languages share the same self-attention parameters. (c) The embeddings of the two languages have their own independent self-attention parameters. For simplicity, we omit parts such as layer normalization, residual connection, linear projection, and soft-max layer, etc.}
		\end{minipage}
		
		
	\end{figure*}
	
	We propose a language-related attention mechanism to reduce multilingual context confusion for E2E code-switching ASR model based on the EC linguistic theory. Our method is based on two view points. First, language switching in code-switching is extremely random. There are almost infinite switching combinations. Trying to cover all combinations through text generation is unrealistic. Second, the EC linguistic theory states that any monolingual fragment that occurs in a code-switching sentence must occurs in one of the monolingual sentences. This linguistic theory establishes a bridge between monolingual data and code-switching data. On this basis, we propose a method to reduce multilingual contextual confusion that with monolingual data. Specifically, our method is based on transformer structure, which has achieved outstanding performance in the field of ASR. This method uses self-attention mechanism to model language context information, as in the field of natural language processing. Due to the randomness of language switching, attention calculations between different languages are difficult and aggravate the confusion between contexts. Therefore, we adjust the attention calculation method between different languages to make the model more emphasis on the context of the same language. This strategy can reduce the complexity of multilingual context and utilize rich monolingual data more efficiently. According to the EC theory, our method theoretically covers most of language contexts. 
	
	Here are our main contributions in this work. First, we propose a language-related attention mechanism to reduce multilingual context confusion for E2E code-switching ASR model based on the EC linguistic theory. Second, we propose three specific attention schemes. Finally, we conduct experiments and demonstrate that our methods is more effective in using monolingual data than the baseline model with ASRU 2019 Mandarin-English code-switching challenge dataset \cite{DBLP:journals/corr/abs-2007-05916}.
	
	The paper is organized as follows. In Section 2, we briefly review the structure of Speech-Transformer ASR model. In Section 3, we introduce the language-related attention strategies of our method in details. In Section 4, we describe our experimental design and discuss the experimental results. Finally, we conclude this paper in Section 5.

	\section{Speech-Transformer}
	\label{sec:pagestyle}
	
	In order to describe our method in a clear manner, we first review the structure of Speech-Transformer model \cite{8462506}. Speech-Transformer is a successful ASR transformer-based model \cite{DBLP:conf/nips/VaswaniSPUJGKP17}, which is composed of an encoder and a decoder.
	
	The encoder encodes and transforms acoustic features to high-level representation. First, we apply a convolutional neural network (CNN) module to down-sample the acoustic feature sequence and obtain an initial hidden representation. Then we apply several layers of encoder blocks of the same structure to further encode the down-sampled acoustic representation. Each encoder block consists of two sub-blocks: multi-head self-attention and position-wise feed-forward layer. Layer normalization and residual connections are applied between sub-blocks to stabilize training process and improve performance.
	
	The decoder receives the output of the encoder and performs a loss calculation by matching the acoustic representation with target text. First, we apply word embedding layer to convert discrete modeling units into vector representations. Then we apply a stack of decoder blocks to encode the text context and interact with the encoded acoustic representation subsequently. The decoder block consists of three parts: masked multi-head self-attention, multi-head cross-attention and position-wise feed-forward layer. The masked multi-head self-attention models language context information, which is the focus of this paper. The queries, keys, and values are the word embedding representation. The multi-head cross-attention completes the information interaction between the decoder and the encoder through the attention mechanism. The rest of the decoder structure is similar to the encoder.
	
	\section{Language-Related Attention Mechanism}
	
	According to the EC linguistic theory, any monolingual fragment that occurs in a code-switching sentence must occur in one of monolingual sentences. This linguistic theory establishes a bridge between monolingual data and code-switching data. General E2E models cannot effectively model such linguistic constraints. Specifically, for transformer model, the multi-head self-attention module of the decoder learns language context information. The module uses the same attention calculation method for different language modeling units. Different languages are treated as the same language. Multilingual contextual information between languages cannot be obtained from monolingual data. Monolingual data sometimes even reduce performance of the code-switching ASR model. Therefore, we modify the self-attention of the decoder to strengthen contextual relationship of the same language and reduce contextual relationship between different languages. Multilingual contextual knowledge can be learned more effectively from rich monolingual data. We have implemented three specific language-related attention schemes. The corresponding model structures are shown in Fig. 1.
	
	As shown in Fig. 1(a), we re-weight the self-attention scores according to the language. Specifically, when the attention scores of a  modeling unit are calculated with respect to other units in the sentence, the attention scores belonging to the same language are increased while the attention scores of different language modeling units are reduced. In this way, a stronger connection is established between the modeling units of the same language. The mutual influence between different languages is suppressed. Therefore, language context representation learned from code-switching data is more compatible with monolingual data. The model can extract context information from monolingual data more effectively.  
	
	The calculation of the attention is expressed as,
	\begin{equation}
	Attention(\mathbf{Q},\mathbf{K},\mathbf{V}) = Softmax(\frac{\mathbf{Q}\mathbf{K^{T}}}{\sqrt{d_k}})\mathbf{V}
	\end{equation}
	where $\mathbf{Q},\mathbf{K},\mathbf{V}$ denote the query, key, and value respectively, $d_k$ is the dimension of the key. For the self-attention, the query, key, and value are the target text embedding. Our method is formally expressed as
	\begin{equation}
	OurAttention(\mathbf{Q},\mathbf{K},\mathbf{V},\mathbf{W}) = Softmax(\frac{\mathbf{Q}\mathbf{K^T}}{\sqrt{d_k}} \mathbf{W})\mathbf{V}
	\end{equation}
	where $\mathbf{W}$ denotes the re-weighting matrix, which is obtained according to the language switching distribution in the sample. The matrix completes the process of calculating language-related attention by adjusting the word embedding matrix or the attention score matrix.
	
	In addition to adjusting the attention scores, we also implement two other different attention methods.  As shown in Fig. 1(b), we separate the code-switching embedding representation by languages and get the embedding sequences of the two languages. Then these two sequences are treated as monolingual cases for attention calculation. In Fig. 1(b), the two sequences share the same attention calculation parameters. After obtaining the corresponding context representations, the representations of the two languages are merged to obtain a complete code-switching text representation. This approach is able to achieve the goal of strengthening the connection with the same language while reducing the interference of different languages. 
	
	The third method described in Fig. 1(c) is similar to the one in Fig. 1(b), except that the two language embedding sequences have their own independent self-attention parameters. The design of this model structure minimizes mutual interference between the two languages. This is a more thorough way of calculating language-related attention and theoretically can make use of monolingual data more effectively. After completing the self-attention calculation, the context fusion of the two languages is the same as Fig. 1(b). 
	Note that the subsequent cross attention calculation is completely consistent with the ordinary Speech-Transformer model.
	
	In the training stage, three implementation schemes mentioned above require the participation of a re-weighting matrix, which is generated according to the target text during the data reading phase. In the test stage, the matrix is generated and expanded in real time with the auto-regressive decoding process. The method does not perform language identification in advance, but only on the basis of the already decoded results.
	

	\section{Experiments}
	\label{sec:typestyle}
	
	\subsection{Datasets}
	\label{ssec:subhead}
	We perform experiments with ASRU 2019 Mandarin-English code-switching challenge dataset.
	The corpus consists of about 200 hours code-switching training data and 500 hours monolingual Mandarin training data \cite{DBLP:journals/corr/abs-2007-05916}.  The audio data are collected by mobile phones in quiet environments with 16kHz sampling rate. To facilitate experimental design, we further include the 460-hour subset of Librispeech English dataset \cite{DBLP:conf/icassp/PanayotovCPK15} into training set. The development set and the test set each consists of 20 hour code-switching data.
	
	\subsection{Experiment Setups}
	\label{ssec:subhead}
	
	We apply 40-dimensional filter-banks with frame length of 25 ms and frame shift of 10 ms as acoustic features. The modeling units for Chinese and English are characters and word pieces respectively. We choose 1.0k English word pieces as the English modeling units. 
	For Chinese, we reserve characters with more than 10 occurrences in the training set as modeling units. There are about 3.1k Chinese characters in total. We apply mix error rate (MER) as the evaluation metric. It counts word error for English and character error for Chinese. In monolingual cases, we compute word error rate (WER) for English and character error rate (CER) for Chinese. 
	The metric is widely adopted to evaluate the Mandarin-English code-switching ASR system.
	
	We implement all the models with Speech-Transformer architecture. We down-sample the acoustic features with two $3\times3$ 2D-CNN layers with stride 2. The dimension of the subsequent linear layer is 512. We apply relative position encoding to model position information. We use a convolution-augmented transformer (conformer) \cite{gulati2020conformer} as the acoustic encoder with dimension 512. The number of attention heads is 4. The size of the convolution kernel is 32. For decoder, the attention dimension is 512. The number of the head is 4. The dimension of position-wise feed-forward networks is 1024. There are 12 and 6 blocks for encoder and decoder respectively. In our method, we need to separate two monolingual embeddings from mixed embeddings. Specifically, by weighting the embedding of one language, we obtain the embedding representation of another language. The weighting parameter is set to 0.1. To further improve the performance of the model, the weighted sum of the connectionist temporal classification (CTC) loss and cross-entropy loss is used as the final loss function. The weighting parameter of CTC loss is set to 0.2. The weight parameter of the CTC score during decoding is set to 0.3.
	
	We apply uniform label smoothing \cite{muller2019does} to avoid over-fitting. The parameter is set to 0.1. We also apply SpecAugment with frequency masking (F=30, mF=2) and time masking (T=40, mT=2) to improve robustness of the models \cite{DBLP:conf/interspeech/ParkCZCZCL19}. We apply residual connections between sub-blocks for stable training \cite{he2016deep}. We set the parameter of residual dropout to 0.1 \cite{srivastava2014dropout}. The model is trained using 2 NVIDIA V100 GPUs with the optimization strategy of $\beta_{1}=0.9, \beta_{2}=0.998, \epsilon=1e^{-8}$ \cite{8462506}. The batch-size is set to 32. The learning rate is set by a warm-up strategy \cite{DBLP:conf/nips/VaswaniSPUJGKP17}. We apply model averaging to improve model performance. During beam-search decoding, the beam-width is set to 10.
	
	\begin{table}[htb]
		\centering
		\caption{MER/CER/WER (\%) of our methods and baseline model with only code-switching data. MER refers to whole utterances. CER refers to Chinese part and WER refers to the English part. `score re-weighted', `attention shared' and `attention not shared' respectively correspond to the aforementioned three self-attention adjustment schemes.} \label{tab:aStrangeTable} %
		\renewcommand\tabcolsep{2.5pt}
		\begin{tabular}{ccccccc}
			\toprule[1.2pt]
			\multirow{2}*{model}& \multicolumn{3}{c}{Dev} & \multicolumn{3}{c}{Test}\\
			\cmidrule(r){2-4} \cmidrule(r){5-7}
			~ & MER & CER & WER & MER & CER & WER \\
			\hline
			baseline & 12.08 & 9.85 & 30.04 & 11.12 & 9.03 & 28.33 \\ 
			\hline
			score re-weighted & 12.15 & 9.93 & 30.03 & 11.21 & 9.11 & 28.52 \\
			attention shared & 11.40 & 9.28 & 28.54 & 10.75 & 8.71 & 27.57 \\
			attention not shared & \textbf{11.09} & 9.00 & 28.01 & \textbf{10.44} & 8.42 & 27.03 \\
			\bottomrule[1.2pt]
		\end{tabular} 
	\end{table}
	
	\begin{table}[ht] 
		\centering 
		\caption{MER/CER/WER (\%) of transformer baseline with extra monolingual data. `200h CS', `500h CH', `460h EN'  respectively refer to code-switching data, Chinese data, English data and `All' corresponds to the above three datasets.} \label{tab:aStrangeTable} %
		\renewcommand\tabcolsep{2.5pt}
		\begin{tabular}{ccccccccc} 
			\toprule[1.2pt]
			\multirow{2}*{model} &\multirow{2}*{Data}& \multicolumn{3}{c}{Dev} & \multicolumn{3}{c}{Test}\\
			\cmidrule(r){3-5} \cmidrule(r){6-8}
			~ & ~ & MER & CER & WER & MER & CER & WER\\
			\hline
			~ & 200h CS & 12.08 & 9.85 & 30.04 & 11.12 & 9.03 & 28.33 \\ 
			
			\multirow{2}*{baseline} & + 500h CH & 11.48 & 9.00 & 31.43 & 10.46 & 8.14 & 29.56 \\
			
			~ & + 460h EN & 11.99 & 10.01 & 28.03 & 11.24 & 9.41 & 26.33 \\
			
			~ & All  & \textbf{11.19} & 8.99 & 28.87 & \textbf{10.34} & 8.32 & 27.00 \\
			\bottomrule[1.2pt]
		\end{tabular} 
	\end{table}
	
	\begin{table}[ht] 
		\centering 
		\caption{MER/CER/WER (\%) of self-attention sharing strategy with extra monolingual data.} \label{tab:aStrangeTable} %
		\renewcommand\tabcolsep{2.5pt}
		\begin{tabular}{cccccccc} 
			\toprule[1.2pt]
			\multirow{2}*{model} & \multirow{2}*{Data}& \multicolumn{3}{c}{Dev} & \multicolumn{3}{c}{Test}\\
			\cmidrule(r){3-5} \cmidrule(r){6-8}
			~ & ~ & MER & CER & WER & MER & CER & WER\\
			\hline
			~ & 200h CS & 11.40 & 9.28 & 28.54 & 10.75 & 8.71 & 27.57 \\ 
			
			attention & + 500h CH & 9.77 & 7.42 & 28.63 & 9.17 & 6.88 & 28.01 \\
			
			shared & + 460h EN & 11.20 & 9.35 & 26.08 & 10.66 & 8.90 & 25.15 \\
			
			~ & All & \textbf{9.63} & 7.51 & 26.74 & \textbf{8.95} & 6.96 & 25.32 \\
			
			\bottomrule[1.2pt]
		\end{tabular} 
	\end{table}
	
	\begin{table}[ht] 
		\centering 
		\caption{MER/CER/WER (\%) of self-attention independence strategy with extra monolingual data.} \label{tab:aStrangeTable} %
		\renewcommand\tabcolsep{2.5pt}
		\scalebox{0.97}{
			\begin{tabular}{cccccccc} 
				\toprule[1.2pt]
				\multirow{2}*{model} & \multirow{2}*{Data}& \multicolumn{3}{c}{Dev} & \multicolumn{3}{c}{Test}\\
				\cmidrule(r){3-5} \cmidrule(r){6-8}
				~ & ~ & MER & CER & WER & MER & CER & WER\\
				\hline
				~ & 200h CS & 11.09 & 9.00 & 28.01 & 10.44 & 8.42 & 27.03 \\ 
				
				attention & + 500h CH & 9.40 & 7.03 & 28.54 & 8.79 & 6.52 & 27.43 \\
				
				not shared & + 460h EN & 11.08 & 9.31 & 25.31 & 10.22 & 8.55 & 24.00 \\
				
				~ & All & \textbf{9.34} & 7.25 & 26.18 & \textbf{8.57} & 6.68 & 24.11 \\
				
				\bottomrule[1.2pt]
			\end{tabular} 
		}
	\end{table}
	
	\subsection{Performance with Code-switching Data}
	\label{ssec:subhead}
	We design a comparative experiment between the proposed method and the baseline model based on the code-switching data. The experimental results are shown in Table 1. The attention scores re-weighting strategy slightly degrades the performance of the model. The performance degradation may be due to the confusion of attention caused by the artificial post-adjustment of the attention scores. The other two attention adjustment strategies improve the performance of the model to a certain extent. Due to the small amount of code-switching data, this experimental result is unable to fully reflect the superiority of our method. 
	The advantage of our method is the ability of using monolingual data. 
	
	\subsection{Performance with Extra Monolingual Data}
	\label{ssec:subhead}
	
	To demonstrate the superiority of the proposed method in utilizing monolingual data, we perform experimental comparisons between the baseline model and our methods with additional monolingual data. The experimental results of the baseline model with extra monolingual data are listed in Table 2. The results show that when we add monolingual data, recognition error rate of the corresponding language is reduced. 
	However, this action leads to negative impact on other language. 
	A possible reason is mutual interference between the two language data. 
	Adding monolingual data does not always improve the MER of code-switching ASR task for the ordinary Speech-Transformer model. The baseline model is relatively inefficient in improving code-switching ASR performance using extra monolingual data. Finally, the performance of the baseline model with all training data achieves 7.01\% relative MER reduction compared with the results of only code-switching data.
	
	Then we conduct experiments on our proposed methods with the same data design scheme. Since the experimental results of the self-attention scores re-weighting strategy are not satisfactory and the limited paper space, we only show the experimental results of the other two methods. Table 3 shows the experimental results of the self-attention sharing strategy. This method achieves greater improvement with extra monolingual data than the baseline model. Compared with the baseline model, the degradation of extra monolingual data to the performance of the competing language is greatly reduced. The results support our claims that our method can reduce context interference between the two languages and improve performance of code-switching ASR model. The results also demonstrate that our method is more effective in using monolingual data. Finally, the performance of the proposed model with all training data achieves 16.28\% relative MER reduction compared with the results trained with only code-switching data. 
	
	Table 4 shows the experimental results of the self-attention independence strategy. The experimental results are consistent to Table 3. The strategy with independent self-attention for each language is more effective and achieves the best recognition performance. Since the two languages carry out context modeling independently, mutual interference further reduces. Overall, the proposed model achieves 17.91\% relative reduction in MER compared with the results of only code-switching data. Compared with the baseline model, the proposed method achieves 17.12\% relative reduction in MER. The above experimental results suggest that our method is an effective solution for code-switching ASR tasks.

	\section{Conclusion}
	\label{sec:majhead}
	
	In this paper, we propose a language-related attention mechanism to reduce multilingual context confusion for the E2E code-switching ASR model based on the EC linguistic theory. These methods can reduce mismatch between code-switching data and monolingual data in modeling language context. Experimental results suggest that the proposed methods utilize monolingual data effectively to improve the performance of code-switching ASR task, compared to the baseline model. 
	\section{Acknowledgment}
	\label{sec:majhead}
	This work is supported by the Key Research Project of China (No.2019KD0AD01), the National Natural Science Foundation of China (NSFC) (No.61901473, No.62101553, No.61831022). This research is funded by Huawei Noah’s Ark Lab.
	\vfill\pagebreak
	


	\bibliographystyle{IEEEtran}
	
	\bibliography{mybib}

\begin{thebibliography}{10}
\providecommand{\url}[1]{#1}
\csname url@samestyle\endcsname
\providecommand{\newblock}{\relax}
\providecommand{\bibinfo}[2]{#2}
\providecommand{\BIBentrySTDinterwordspacing}{\spaceskip=0pt\relax}
\providecommand{\BIBentryALTinterwordstretchfactor}{4}
\providecommand{\BIBentryALTinterwordspacing}{\spaceskip=\fontdimen2\font plus
\BIBentryALTinterwordstretchfactor\fontdimen3\font minus
  \fontdimen4\font\relax}
\providecommand{\BIBforeignlanguage}[2]{{%
\expandafter\ifx\csname l@#1\endcsname\relax
\typeout{** WARNING: IEEEtran.bst: No hyphenation pattern has been}%
\typeout{** loaded for the language `#1'. Using the pattern for}%
\typeout{** the default language instead.}%
\else
\language=\csname l@#1\endcsname
\fi
#2}}
\providecommand{\BIBdecl}{\relax}
\BIBdecl

\bibitem{muysken2000bilingual}
P.~Muysken, P.~C. Muysken \emph{et~al.}, \emph{Bilingual speech: A typology of
  code-mixing}.\hskip 1em plus 0.5em minus 0.4em\relax Cambridge University
  Press, 2000.

\bibitem{graves2006connectionist}
A.~Graves, S.~Fern{\'a}ndez, F.~Gomez, and J.~Schmidhuber, ``Connectionist
  temporal classification: labelling unsegmented sequence data with recurrent
  neural networks,'' in \emph{Proceedings of the 23rd international conference
  on Machine learning}.\hskip 1em plus 0.5em minus 0.4em\relax ACM, 2006, pp.
  369--376.

\bibitem{graves2013speech}
A.~Graves, A.-r. Mohamed, and G.~Hinton, ``Speech recognition with deep
  recurrent neural networks,'' in \emph{2013 IEEE international conference on
  acoustics, speech and signal processing}.\hskip 1em plus 0.5em minus
  0.4em\relax IEEE, 2013, pp. 6645--6649.

\bibitem{chan2016listen}
W.~Chan, N.~Jaitly, Q.~Le, and O.~Vinyals, ``Listen, attend and spell: A neural
  network for large vocabulary conversational speech recognition,'' in
  \emph{2016 IEEE International Conference on Acoustics, Speech and Signal
  Processing (ICASSP)}.\hskip 1em plus 0.5em minus 0.4em\relax IEEE, 2016, pp.
  4960--4964.

\bibitem{8462506}
L.~{Dong}, S.~{Xu}, and B.~{Xu}, ``Speech-transformer: A no-recurrence
  sequence-to-sequence model for speech recognition,'' in \emph{2018 IEEE
  International Conference on Acoustics, Speech and Signal Processing
  (ICASSP)}, 2018, pp. 5884--5888.

\bibitem{rao2017exploring}
K.~Rao, H.~Sak, and R.~Prabhavalkar, ``Exploring architectures, data and units
  for streaming end-to-end speech recognition with rnn-transducer,'' in
  \emph{2017 IEEE Automatic Speech Recognition and Understanding Workshop
  (ASRU)}.\hskip 1em plus 0.5em minus 0.4em\relax IEEE, 2017, pp. 193--199.

\bibitem{gulati2020conformer}
A.~Gulati, J.~Qin, C.-C. Chiu, N.~Parmar, Y.~Zhang, J.~Yu, W.~Han, S.~Wang,
  Z.~Zhang, Y.~Wu \emph{et~al.}, ``Conformer: Convolution-augmented transformer
  for speech recognition,'' \emph{Proc. Interspeech 2020}, pp. 5036--5040,
  2020.

\bibitem{li2019towards}
K.~Li, J.~Li, G.~Ye, R.~Zhao, and Y.~Gong, ``Towards code-switching asr for
  end-to-end ctc models,'' in \emph{ICASSP 2019-2019 IEEE International
  Conference on Acoustics, Speech and Signal Processing (ICASSP)}.\hskip 1em
  plus 0.5em minus 0.4em\relax IEEE, 2019, pp. 6076--6080.

\bibitem{8682674}
B.~{Li}, Y.~{Zhang}, T.~{Sainath}, Y.~{Wu}, and W.~{Chan}, ``Bytes are all you
  need: End-to-end multilingual speech recognition and synthesis with bytes,''
  in \emph{ICASSP 2019 - 2019 IEEE International Conference on Acoustics,
  Speech and Signal Processing (ICASSP)}, 2019, pp. 5621--5625.

\bibitem{8462201}
S.~{Kim} and M.~L. {Seltzer}, ``Towards language-universal end-to-end speech
  recognition,'' in \emph{2018 IEEE International Conference on Acoustics,
  Speech and Signal Processing (ICASSP)}, 2018, pp. 4914--4918.

\bibitem{zhang2021decouple}
S.~Zhang, J.~Yi, Z.~Tian, Y.~Bai, J.~Tao, and Z.~Wen, ``Decoupling
  pronunciation and language for end-to-end code-switching automatic speech
  recognition,'' in \emph{ICASSP 2021 - 2021 IEEE International Conference on
  Acoustics, Speech and Signal Processing (ICASSP)}, 2021, pp. 6249--6253.

\bibitem{DBLP:conf/interspeech/YilmazHL18}
E.~Yilmaz, H.~van~den Heuvel, and D.~A. van Leeuwen, ``Acoustic and textual
  data augmentation for improved {ASR} of code-switching speech,'' in
  \emph{Interspeech 2018}, B.~Yegnanarayana, Ed.\hskip 1em plus 0.5em minus
  0.4em\relax {ISCA}, 2018, pp. 1933--1937.

\bibitem{DBLP:conf/interspeech/ChangCL19}
C.~Chang, S.~Chuang, and H.~Lee, ``Code-switching sentence generation by
  generative adversarial networks and its application to data augmentation,''
  in \emph{Interspeech 2019}, G.~Kubin and Z.~Kacic, Eds.\hskip 1em plus 0.5em
  minus 0.4em\relax {ISCA}, 2019, pp. 554--558.

\bibitem{li2012code}
Y.~Li and P.~Fung, ``Code-switch language model with inversion constraints for
  mixed language speech recognition,'' in \emph{Proceedings of COLING 2012},
  2012, pp. 1671--1680.

\bibitem{li2013improved}
Y.~Li and P.~Fung, ``Improved mixed language speech recognition using
  asymmetric acoustic model and language model with code-switch inversion
  constraints,'' in \emph{2013 IEEE International Conference on Acoustics,
  Speech and Signal Processing}.\hskip 1em plus 0.5em minus 0.4em\relax IEEE,
  2013, pp. 7368--7372.

\bibitem{li2014language}
Y.~Li and P.~Fung, ``Language modeling with functional head constraint for code
  switching speech recognition,'' in \emph{Proceedings of the 2014 Conference
  on Empirical Methods in Natural Language Processing (EMNLP)}, 2014, pp.
  907--916.

\bibitem{joshi1982processing}
A.~Joshi, ``Processing of sentences with intra-sentential code-switching,'' in
  \emph{Coling 1982: Proceedings of the Ninth International Conference on
  Computational Linguistics}, 1982.

\bibitem{pfaff1979constraints}
C.~W. Pfaff, ``Constraints on language mixing: Intrasentential code-switching
  and borrowing in spanish/english,'' \emph{Language}, pp. 291--318, 1979.

\bibitem{bhatt1995code}
R.~M. Bhatt, ``Code-switching and the functional head constraint,'' in
  \emph{Janet Fuller et al. Proceedings of the Eleventh Eastern States
  Conference on Linguistics. Ithaca, NY: Department of Modern Languages and
  Linguistics}, 1995, pp. 1--12.

\bibitem{DBLP:conf/interspeech/SriramJSC18}
A.~Sriram, H.~Jun, S.~Satheesh, and A.~Coates, ``Cold fusion: Training seq2seq
  models together with language models,'' in \emph{Interspeech 2018, 19th
  Annual Conference of the International Speech Communication Association},
  B.~Yegnanarayana, Ed.\hskip 1em plus 0.5em minus 0.4em\relax {ISCA}, 2018,
  pp. 387--391.

\bibitem{DBLP:conf/interspeech/ZhaoSRRBLP19}
D.~Zhao, T.~N. Sainath, D.~Rybach, P.~Rondon, D.~Bhatia, B.~Li, and R.~Pang,
  ``Shallow-fusion end-to-end contextual biasing,'' in \emph{Interspeech 2019,
  20th Annual Conference of the International Speech Communication
  Association}, G.~Kubin and Z.~Kacic, Eds.\hskip 1em plus 0.5em minus
  0.4em\relax {ISCA}, 2019, pp. 1418--1422.

\bibitem{DBLP:conf/interspeech/BaiYTTW19}
Y.~Bai, J.~Yi, J.~Tao, Z.~Tian, and Z.~Wen, ``Learn spelling from teachers:
  Transferring knowledge from language models to sequence-to-sequence speech
  recognition,'' in \emph{Interspeech 2019, 20th Annual Conference of the
  International Speech Communication Association}, G.~Kubin and Z.~Kacic,
  Eds.\hskip 1em plus 0.5em minus 0.4em\relax {ISCA}, 2019, pp. 3795--3799.

\bibitem{DBLP:journals/corr/abs-2007-05916}
X.~Shi, Q.~Feng, and L.~Xie, ``The {ASRU} 2019 mandarin-english code-switching
  speech recognition challenge: Open datasets, tracks, methods and results,''
  \emph{CoRR}, vol. abs/2007.05916, 2020.

\bibitem{DBLP:conf/nips/VaswaniSPUJGKP17}
A.~Vaswani, N.~Shazeer, N.~Parmar, J.~Uszkoreit, L.~Jones, A.~N. Gomez,
  L.~Kaiser, and I.~Polosukhin, ``Attention is all you need,'' in
  \emph{Advances in Neural Information Processing Systems 30: Annual Conference
  on Neural Information Processing Systems}, I.~Guyon, U.~von Luxburg,
  S.~Bengio, H.~M. Wallach, R.~Fergus, S.~V.~N. Vishwanathan, and R.~Garnett,
  Eds., 2017, pp. 5998--6008.

\bibitem{DBLP:conf/icassp/PanayotovCPK15}
V.~Panayotov, G.~Chen, D.~Povey, and S.~Khudanpur, ``Librispeech: An {ASR}
  corpus based on public domain audio books,'' in \emph{2015 {IEEE}, {ICASSP}
  2015}.\hskip 1em plus 0.5em minus 0.4em\relax {IEEE}, 2015, pp. 5206--5210.

\bibitem{muller2019does}
R.~M{\"u}ller, S.~Kornblith, and G.~E. Hinton, ``When does label smoothing
  help?'' \emph{Advances in neural information processing systems}, vol.~32,
  2019.

\bibitem{DBLP:conf/interspeech/ParkCZCZCL19}
D.~S. Park, W.~Chan, Y.~Zhang, C.~Chiu, B.~Zoph, E.~D. Cubuk, and Q.~V. Le,
  ``Specaugment: {A} simple data augmentation method for automatic speech
  recognition,'' in \emph{Interspeech 2019, 20th Annual Conference of the
  International Speech Communication Association}, G.~Kubin and Z.~Kacic,
  Eds.\hskip 1em plus 0.5em minus 0.4em\relax {ISCA}, 2019, pp. 2613--2617.

\bibitem{he2016deep}
K.~He, X.~Zhang, S.~Ren, and J.~Sun, ``Deep residual learning for image
  recognition,'' in \emph{Proceedings of the IEEE conference on computer vision
  and pattern recognition}, 2016, pp. 770--778.

\bibitem{srivastava2014dropout}
N.~Srivastava, G.~Hinton, A.~Krizhevsky, I.~Sutskever, and R.~Salakhutdinov,
  ``Dropout: a simple way to prevent neural networks from overfitting,''
  \emph{The journal of machine learning research}, vol.~15, no.~1, pp.
  1929--1958, 2014.

\end{thebibliography}
	
	
\end{document}